\documentclass[runningheads]{llncs}

\usepackage[utf8]{inputenc} 
\usepackage{graphicx} 
\usepackage{multirow}
\usepackage{adjustbox}
\usepackage{float}
\usepackage{array}
\usepackage{xurl}
\usepackage{booktabs}   
\usepackage{tabularx}
\usepackage{rotating}
\usepackage{colortbl}
\usepackage{amsmath}
\usepackage{amssymb}
\usepackage{url}
\usepackage{bbding}
\usepackage[table]{xcolor}
\providecommand{\orcidID}[1]{}
\usepackage{hyperref}
\begin{document}

\title{PaP-NF: Probabilistic Long-Term Time Series Forecasting via Prefix-as-Prompt Reprogramming and Normalizing Flows}
\titlerunning{PaP-NF: Probabilistic Long-Term Time Series Forecasting}
\author{Minju Kim\orcidID{0009-0004-7965-6252} \and
Youngbum Hur\orcidID{0000-0002-1113-1730}\,\Envelope}

\authorrunning{M. Kim and Y. Hur}

\institute{Department of Industrial Engineering, Inha University, Incheon, Republic of Korea\\
\email{12213093kmj@inha.edu}, 
\email{youngbum.hur@inha.ac.kr}}

\maketitle              
\pagestyle{headings}

\noindent\textbf{Abstract.}
Time series forecasting plays a central role in many real-world applications and has been studied extensively. Most existing approaches have used deterministic models. However, real-world environments exhibit inherently uncertain and complex future behaviors, making single-point predictions insufficient. This highlights the need for probabilistic forecasting that can quantify and represent uncertainty. In this work, we propose PaP-NF, a probabilistic forecasting framework that aligns continuous time series with a frozen LLM using a Prefix-as-Prompt mechanism, and conditions a normalizing flow decoder on the global context extracted by the LLM. The quality of the resulting predictive distributions is evaluated using the Continuous Ranked Probability Score (CRPS), a standard metric in probabilistic forecasting. Across a variety of long-term forecasting benchmarks, PaP-NF captures multi-modal uncertainty robustly while maintaining competitive point-forecast accuracy.
Our official code is available at: \url{https://github.com/democracy04/PaP-NF}

\par          
\vspace{3mm}  


\section{Introduction}

Long-term time series forecasting is critical for various fields, including energy management, healthcare, and traffic control \cite{ComprehensiveSurvey2025}. Compared to short-term tasks, predicting long horizons demands a sophisticated approach to modeling extended temporal dependencies and uncertain future trajectories \cite{Hyndman2018}. Notably, real-world scenarios often involve multiple plausible evolution paths that cannot be fully represented by deterministic outcomes. Most established forecasting methods rely on deterministic regression objectives such as Mean Squared Error (MSE), which produce smooth forecasts that may not fully capture multi-modal behaviors or extreme events. Furthermore, recent architectures leveraging patch-level representations focus on local numerical patterns within confined windows. This localized focus inherently overlooks global temporal contexts essential for understanding complex evolution paths such as distribution shifts and long-term trend transitions. Meanwhile, Large Language Models (LLMs) have emerged as promising candidates for time series forecasting, leveraging their exceptional contextual reasoning capabilities. Yet, a fundamental challenge remains: most LLM-based approaches rely on tokenization, which fragments continuous values and inherently degrades numerical precision. Moreover, using LLMs as direct predictors often provides no principled mechanism for continuous density estimation. This makes it difficult to quantify uncertainty accurately, leaving the capture of diverse, multi-modal future trajectories as a significant open challenge.

\noindent In this paper, we propose PaP-NF, a novel probabilistic forecasting framework designed to address these limitations by utilizing pre-trained LLMs exclusively as global context encoders. Our approach introduces a principled architectural separation between global semantic extraction and local probabilistic generation. This decoupled design enables the model to exploit the high-level reasoning of LLMs without sacrificing the numerical accuracy of continuous time series. By bridging the gap between deterministic predictors and generative probabilistic models, PaP-NF facilitates the estimation of flexible, multi-modal predictive distributions. The major contributions of this work are summarized as follows:
\begin{itemize}
    \item \textbf{Principled hybrid framework:} We propose a unified architecture that preserves local numerical precision via linear embeddings while utilizing frozen LLMs for global semantic reasoning. This hybrid design alleviates the discretization issues of LLM-only predictors.
    \item \textbf{Prefix-based alignment with frozen LLMs:} We introduce a Prefix-as-Prompt (PaP) reprogramming mechanism that aligns numerical embeddings with pre-trained LLMs, enabling global context extraction without modifying LLM parameters.
    \item \textbf{Uncertainty-aware long-horizon prediction:} PaP-NF conditions normalizing flows on joint numerical and LLM contexts, effectively synergizing global context extraction with flexible distribution modeling. This integrated approach allows for robust, multi-modal uncertainty modeling beyond deterministic limitations.
\end{itemize}

\begin{sloppypar}
\section{Related Work}

\textbf{Deep Learning Models for Long-Term Time Series Forecasting.} Early studies on time series forecasting relied on recurrent architectures such as RNNs and LSTMs, while Transformer-based models later enabled more effective modeling of long-range temporal dependencies. Informer \cite{Informer} introduced sparse self-attention for efficient long-term forecasting, followed by variants such as FEDformer \cite{FEDformer} and Autoformer \cite{Autoformer} which leverage frequency domain analysis and decomposition mechanisms. More recently, TimesNet \cite{TimesNet} captures multi-scale temporal variations via 2D tensor transformations, and ETSformer \cite{ETSformer} integrates exponential smoothing into Transformers. Despite their success in modeling temporal patterns, these methods predominantly focus on point forecasting, leaving predictive uncertainty largely unaddressed.
\par\vspace{\baselineskip}
\noindent\textbf{Time Series Analysis with Large Language Models.} Recent studies have explored leveraging the contextual reasoning capabilities of LLMs for the time series domain, although there are ongoing discussions regarding their actual utility in forecasting tasks \cite{AreLLMUsefulTS}. GPT4TS \cite{GPT4TS} adopts a parameter-efficient strategy by training lightweight adapters, while Lag-LLaMA \cite{LagLLaMA} directly uses LLMs as predictors via probabilistic token sampling. Although these approaches demonstrate the potential of LLMs for forecasting, they face inherent limitations in representing continuous numerical signals. In particular, converting real-valued time series into discrete token or prompt representations can degrade fine-grained numerical fidelity, reflecting information loss introduced by discretization rather than insufficient model capacity. This line of work is closely related to parameter-efficient prompt tuning and prefix-tuning methods in NLP \cite{PromptTuning,PrefixTuning}.
\par\vspace{\baselineskip} 
\noindent \textbf{Probabilistic Time Series Forecasting.} Probabilistic forecasting aims to characterize the uncertainty of future trajectories. DeepAR \cite{DeepAR} employs an autoregressive framework with fixed parametric assumptions, such as Gaussian distributions. In contrast, Normalizing Flows \cite{RealNVP,ProbTSNF} provide greater flexibility via invertible transformations for flexible distribution modeling. While diffusion-based models \cite{CSDI} have demonstrated high generative fidelity, their iterative sampling process is often expensive for long horizons. Normalizing Flows offer an efficient alternative, enabling non-iterative sampling while maintaining high model expressivity. 
\par\vspace{\baselineskip} 
\noindent\textbf{Positioning of PaP-NF.} Existing prefix-based methods like Time-LLM \cite{TimeLLM} typically utilize the LLM as the primary forecasting backbone. In contrast, PaP-NF adopts a decoupled architecture: the LLM serves exclusively as a global context encoder via prefix-based alignment, while the generative task is offloaded to a conditional Normalizing Flow \cite{NFReview}. This separation leverages the semantic reasoning of LLMs alongside the continuous density estimation of flows. Consequently, PaP-NF circumvents the discretization artifacts of LLM-centric predictors and the sampling latency of diffusion-based frameworks.
\end{sloppypar}

\section{Methodology} 
\begin{figure}
    \centering
    \includegraphics[width=1\linewidth]{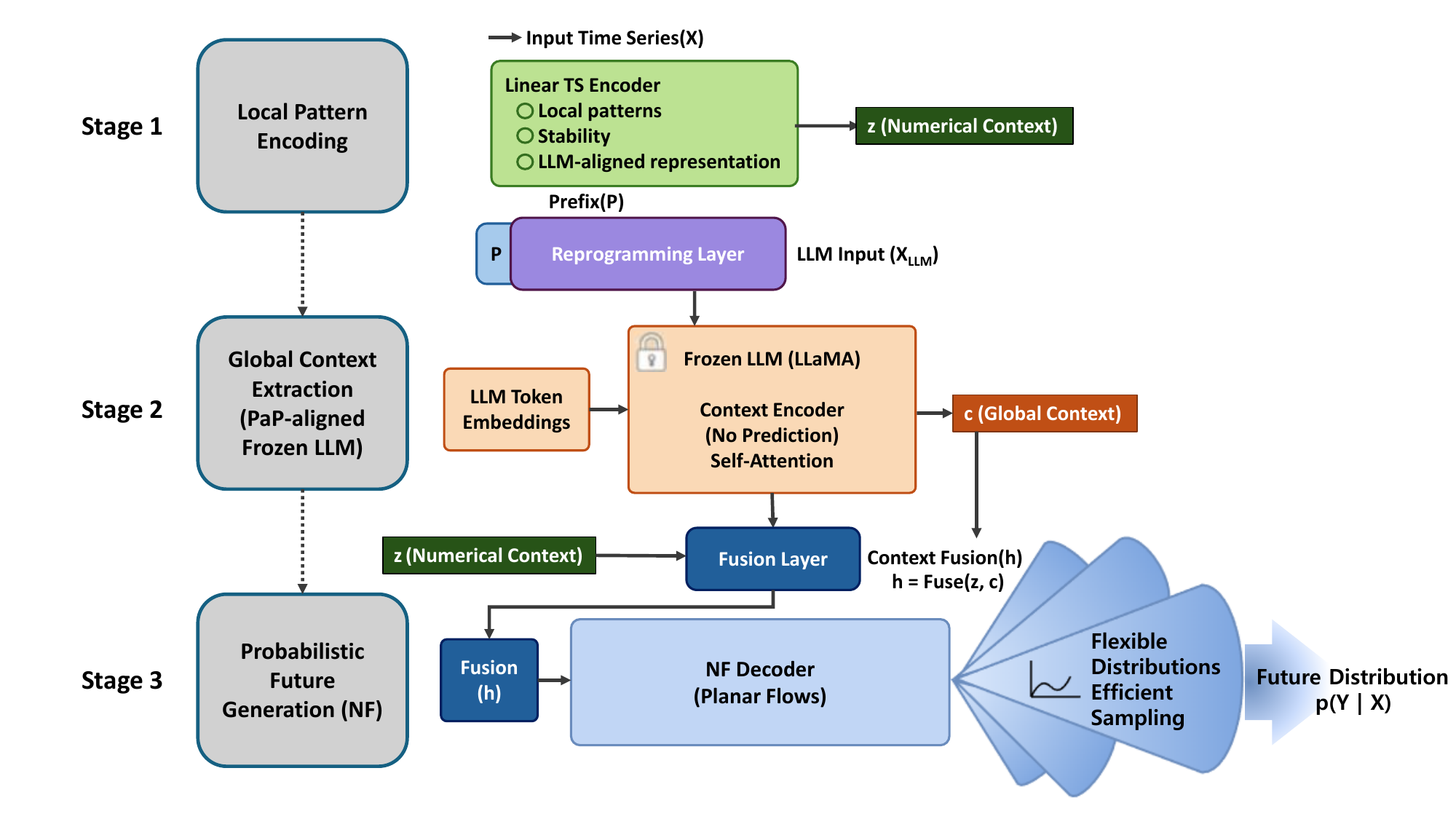}
    \caption{Overview of the PaP-NF framework. Given input time series $X$, a linear encoder extracts localized temporal patterns as $\mathbf{z}$. Learnable prefixes $\mathbf{P}$ align $\mathbf{z}$ with a frozen LLM, which produces a global context vector $\mathbf{c}$ via average pooling. The fused representation $\mathbf{h} =\mathrm{Fuse}(\mathbf{z}, \mathbf{c})$ conditions a normalizing flow to generate the forecast distribution.}
    \label{fig:Overview}
\end{figure}

PaP-NF combines a lightweight numerical encoder with a frozen LLM that provides high-level contextual reasoning. The numerical component captures localized temporal variations, while the LLM contributes broader semantic structure. These two representations are integrated within a unified probabilistic framework. Unlike existing LLM-based forecasters that force the model to produce numerical predictions directly, we use the LLM exclusively as a context encoder. To model predictive uncertainty, we employ a conditional normalizing flow. This module receives both the numerical features and the global context extracted by the LLM, and generates a full predictive distribution. The overall architecture and stage-wise data flow of PaP-NF are summarized in Fig.\ref{fig:Overview}.

\subsection{Problem Formulation}
Given an input time series $X=\{x_1,\ldots,x_L\}\in\mathbb{R}^{L\times C}$, our goal is to estimate the conditional distribution $p(Y\mid X)$ of a future multivariate time series $Y=\{y_1,\ldots,y_H\}\in\mathbb{R}^{H\times C}$ over the next $H$ steps, where $L$ denotes the look-back window length and $C$ is the number of variables. Whereas conventional regression-based approaches output a single point estimate $\hat{Y}$, real-world future trajectories typically admit multiple plausible outcomes, which calls for explicit probabilistic modeling. To achieve this, the proposed framework extracts two complementary representations from the input time series. First, $f_{\mathrm{num}}(\cdot)$ denotes a numerical encoder that transforms the input $X$ into a numerical representation $\mathbf{z}\in\mathbb{R}^{d_n}$ via linear embedding, where $d_n$ is the dimension of the numerical embedding vector and $d$ denotes the token embedding dimension of the frozen LLM used in subsequent alignment. Second, $f_{\mathrm{glob}}(\cdot)$ extracts a global context representation $\mathbf{c}\in\mathbb{R}^{d_c}$, where $d_c$ is a fixed context dimension obtained by linearly projecting the LLM hidden states. The Prefix-as-Prompt reprogrammed input is passed through the frozen LLM, and the resulting hidden states are aggregated via average pooling to form $\mathbf{c}$.

\noindent When these two representations are combined, the conditional distribution of the future time series can be expressed as:
\begin{equation}
p(Y\mid X)=p\big(Y\mid \mathbf{z},\mathbf{c}\big).
\end{equation}
By explicitly modeling local numerical information and global contextual information separately and integrating them only at the probabilistic generation stage, the proposed framework aims to alleviate representational bottlenecks that often arise in long-term forecasting. As illustrated in Fig.~\ref{fig:Overview}, the proposed framework consists of three stages: numerical encoding for local pattern representation, Prefix-as-Prompt-based global context extraction using a frozen LLM, and conditional probabilistic generation via normalizing flows.

\subsection{Numerical Temporal Encoding and Reprogramming}

\begin{figure}
    \centering
    \includegraphics[width=1\linewidth]{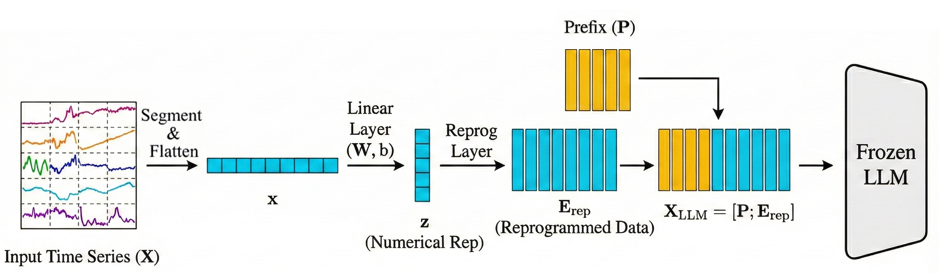}
    \caption{Detailed illustration of the temporal encoding and prompt-based reprogramming process. The input time series $X$ is partitioned into segments and flattened into a numerical vector $\mathbf{z}$ via a linear layer, projected into the LLM token embedding space to obtain $\mathbf{E}_{\mathrm{rep}}$, and concatenated with learnable prefix vectors $\mathbf{P}$ to form the aligned input $X_{\mathrm{LLM}}$ for the frozen LLM.}
    \label{fig:reprogramming}
\end{figure}

\noindent To represent the input time series, we adopt a linear embedding instead of a deep temporal encoder. Motivated by recent findings that linear mappings can be competitive with more complex architectures for forecasting \cite{DLinear}, this choice reduces overfitting risk and alleviates structural mismatch between numerical time series representations and the LLM embedding space. As illustrated in Fig.\ref{fig:reprogramming}, the input sequence $X \in \mathbb{R}^{L \times C}$ is first partitioned into localized segments. Each segment is then flattened and concatenated to form the vectorized representation  $\mathbf{x}\in\mathbb{R}^{LC}$, where $LC$ is the flattened dimension ($L \times C$). A linear transformation is then applied to generate the numerical representation $\mathbf{z}$:
\begin{equation}
\mathbf{z}=W\mathbf{x}+b,
\end{equation}
where $W\in\mathbb{R}^{d_n\times LC}$ and $b\in\mathbb{R}^{d_n}$ are learnable parameters. This linear transformation produces a compact numerical representation $\mathbf{z}$ that summarizes local temporal variations and inter-variable correlations.

\noindent To align $\mathbf{z}$ with the LLM input space, we apply a lightweight reprogramming layer
that projects $\mathbf{z}$ into the $d$-dimensional token embedding space of the frozen LLM,
producing a projected vector $\mathbf{e} \in \mathbb{R}^{d}$:
\begin{equation}
\mathbf{e} = W_p \mathbf{z} + b_p.
\end{equation}

\noindent In practice, the input sequence is partitioned into $M$ patches, and each patch is projected into the LLM embedding space, forming the reprogrammed token sequence $\mathbf{E}_{\mathrm{rep}} \in \mathbb{R}^{M \times d}$.
We then prepend a learnable prefix
matrix $\mathbf{P} \in \mathbb{R}^{K \times d}$ to $\mathbf{E}_{\mathrm{rep}}$, where $K$
denotes the prefix length. The aligned LLM input is finally constructed as:
\begin{equation}
X_{\mathrm{LLM}} = [\mathbf{P}; \mathbf{E}_{\mathrm{rep}}].
\end{equation}

\noindent This mapping aligns the local temporal features with the semantic space of the frozen LLM, completing the numerical encoding process $\mathbf{z} = f_{\mathrm{num}}(X)$ introduced
in Section~3.1.

\subsection{Global Context Modeling with Frozen LLM}
In PaP-NF, the frozen LLM functions as a semantic pattern encoder that abstracts high-level temporal structures, rather than a direct numerical predictor. The linear encoder retains local numerical structure but is not sufficient for capturing long-range temporal dependencies or regime-level variations. The LLM compensates for this limitation by modeling global temporal semantics. By aligning numerical representations with the LLM embedding space via Prefix-as-Prompt reprogramming, the LLM’s self-attention aggregates these representations into a global context vector that reflects the semantic organization of temporal dynamics and complements local numerical features.

\noindent The aligned input $X_{\mathrm{LLM}}$ is processed by the frozen LLM, producing contextualized hidden representations for the reprogrammed tokens.
The LLM’s self-attention mechanism integrates global interactions across the entire sequence, producing token-wise hidden states $\{h_n\}_{n=1}^{N} \in \mathbb{R}^{d}$. To obtain the global context representation $\mathbf{c} \in \mathbb{R}^{d_c}$, these hidden states are projected into the context dimension $d_c$ using a linear layer $W_c \in \mathbb{R}^{d_c \times d}$ with bias $b_c \in \mathbb{R}^{d_c}$, and then aggregated via average pooling. This projection-plus-pooling strategy provides a simple and robust summarization of long-term trends and high-level temporal semantics without introducing excessive learnable parameters. Formally, the global context extractor $f_{\mathrm{glob}}(\cdot)$ is defined as:

\begin{equation}
\mathbf{c} = f_{\mathrm{glob}}(X)
= \frac{1}{N}\sum_{n=1}^{N} \bigl(W_c h_n + b_c\bigr),
\end{equation}
where $h_n \in \mathbb{R}^{d}$ denotes the hidden state of the $n$-th token from the frozen LLM, $W_c \in \mathbb{R}^{d_c \times d}$ and $b_c \in \mathbb{R}^{d_c}$ are learnable projection parameters, and $N = K + M$ denotes the total sequence length of the aligned input $X_{\mathrm{LLM}}$. This summary vector $\mathbf{c} \in \mathbb{R}^{d_c}$ serves as a global context encapsulating long-term trends and high-level interpretations of local patterns, and is subsequently used as a conditioning signal in the probabilistic generation stage.

\subsection{Conditional Normalizing Flow for Probabilistic Forecasting}

\begin{figure}[H]
    \centering
    \includegraphics[width=0.95\linewidth]{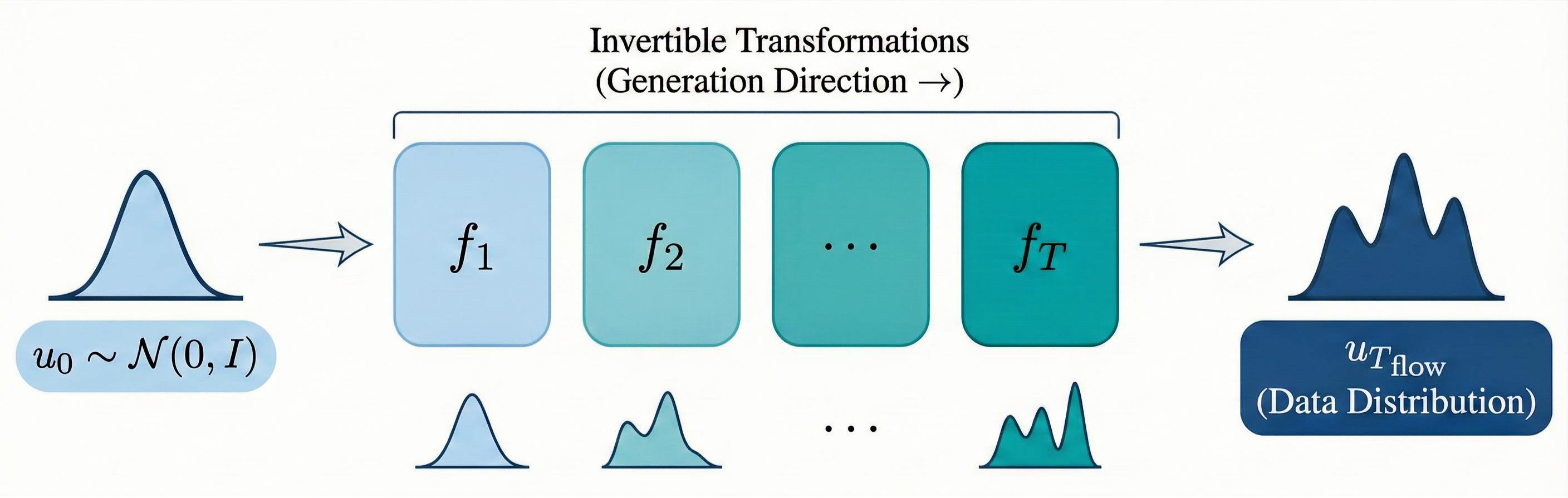}
    \caption{Illustration of the normalizing flow framework. A simple base distribution $u_0 \sim \mathcal{N}(0, I)$ is transformed into a complex target distribution $u_{T_{\mathrm{flow}}}$ through a sequence of invertible mappings $f_1,\dots,f_T$, enabling flexible distribution modeling and efficient sampling of diverse future trajectories.}
    \label{fig:nf}
\vspace{-5mm}
\end{figure}

In the final stage, a conditional normalizing flow generates the predictive distribution of the future time series using a condition vector $\mathbf{h}$ that combines the local numerical representation $\mathbf{z}$ and the global context representation $\mathbf{c}$. The condition vector is defined as:
\begin{equation}
\mathbf{h}=\mathrm{Fuse}(\mathbf{z},\mathbf{c}),
\end{equation}
where $\mathrm{Fuse}(\cdot)$ denotes a fusion operator that integrates local and global features through concatenation and linear projection to a $d_h$-dimensional space. In our implementation, we concatenate $\mathbf{z}$ and $\mathbf{c}$ and apply a linear projection, yielding $\mathbf{h} = W_h[\mathbf{z}; \mathbf{c}] + b_h$,
where $W_h \in \mathbb{R}^{d_h \times (d_n + d_c)}$ and $b_h \in \mathbb{R}^{d_h}$ are learnable parameters. The overall framework of the normalizing flow decoder is illustrated in Fig.~\ref{fig:nf}. Instead of directly mapping to the data space via a likelihood-based formulation, we employ a latent-variable formulation in which a Gaussian latent variable $u_0 \sim \mathcal{N}(0, I)$ is first sampled and then transformed through a sequence of planar flow layers conditioned on $\mathbf{h}$:

\begin{equation}
u_{T_{\mathrm{flow}}}
=
f_{T_{\mathrm{flow}}}
\circ\cdots\circ
f_1(u_0;\mathbf{h}),
\end{equation}

\noindent 
where $f_t(\cdot)$ denotes the $t$-th planar flow transformation. The transformed latent representation $u_{T_{\mathrm{flow}}}$ is then concatenated with the conditioning vector $\mathbf{h}$ and passed through a reconstruction network to generate the future sample:

\begin{equation}
\hat{Y} = g([u_{T_{\mathrm{flow}}}; \mathbf{h}]).
\end{equation}

\noindent 
This formulation enables efficient stochastic generation of multiple future trajectories, facilitating uncertainty-aware forecasting.

\section{Experiments}

We evaluate the PaP-NF framework through: (i) point forecasting comparisons on long-term benchmarks, (ii) probabilistic performance analysis via density estimation, and (iii) ablation studies to verify the contribution of each component. Our evaluation follows the standard protocols and data split strategies established in prior long-term forecasting literature \cite{TimeLLM,DLinear}. To ensure a fair comparison, all input lengths and prediction horizons are kept identical to those in baseline studies.

\subsection{Experimental Setup}
\begin{sloppypar}
We evaluate PaP-NF on the ETT (ETTh1, ETTh2, ETTm1, ETTm2) and Traffic benchmarks with horizons $H \in \{96, 192, 336, 720\}$. For each dataset, we select hyperparameters by grid-searching look-back windows $L \in \{96, 192, 336, 720\}$, batch sizes in $B \in [1, 16]$, and learning rates in $[10^{-5}, 10^{-3}]$. The model uses a frozen Meta-Llama-3.1 backbone \cite{MetaLlama3,LLaMA} and is trained for 15 epochs on an NVIDIA RTX A5000. Hyperparameters are selected based on validation performance. Additionally, to assess probabilistic accuracy, we report weighted CRPS at a short horizon ($H=24$). We adopt the 24-step horizon to align with established practice in probabilistic forecasting literature \cite{LagLLaMA}, which typically evaluates distributional accuracy at short horizons. CRPS is estimated from 100 stochastic samples drawn from the normalizing flow at each time step. Probabilistic baselines use the standard AutoGluon \cite{GluonTS} implementations to ensure consistent model configurations across datasets. During training, the model is optimized using a reconstruction objective between the predicted output and the ground truth future sequence. At inference time, multiple stochastic samples are generated by repeatedly sampling from the latent space and passing through the flow-based decoder.

\end{sloppypar}

\subsection{Main Results: Point Forecasting Performance}

Table~\ref{tab:long_term_forecast_side_dataset} compares the point forecasting performance of PaP-NF against state-of-the-art baselines. PaP-NF consistently achieves competitive results across all benchmarks and maintains robustness even as the prediction horizon $H$ increases. Notably, on the ETTh2 and ETTm2 datasets with $H=720$, PaP-NF outperforms TimesNet—a strong Transformer-based competitor—reducing MSE by 2.4\% and 3.2\%, respectively. Furthermore, PaP-NF yields the lowest MSE across all horizons on the Traffic dataset, highlighting its stability in modeling high-dimensional volatility. Overall, these results indicate that PaP-NF effectively mitigates error accumulation in long-term forecasting, particularly as the prediction horizon extends.

\begin{table}[!ht]
\centering
\caption{Long-term forecasting performance comparison with major baselines ($\mathbf{H \in \{96, 192, 336, 720\}}$). Results are reported as \textbf{MSE}/\textbf{MAE}, with lower values indicating better performance. The best performance is shown in \textbf{bold}.}

\vspace{5pt}
\label{tab:long_term_forecast_side_dataset}
\renewcommand{\arraystretch}{1.1}
\setlength{\tabcolsep}{3pt}
\small
\begin{tabular}{c|c|cccc}
\hline
\multirow{2}{*}{Dataset} & \multirow{2}{*}{Model} & \multicolumn{4}{c}{Prediction Horizon ($H$)} \\
\cline{3-6}
& & {96} & {192} & {336} & {720} \\
\hline \hline

\multirow{7}{*}{ETTh1}
& Autoformer \cite{Autoformer} & 0.449/0.459 & 0.500/0.482 & 0.521/0.496 & 0.514/0.512 \\
& FEDformer \cite{FEDformer}   & 0.376/0.419 & \textbf{0.420}/0.448 & 0.459/0.465 & 0.506/0.507 \\
& Stationary \cite{Stationary} & 0.513/0.491 & 0.534/0.504 & 0.588/0.535 & 0.643/0.616 \\
& ETSformer \cite{ETSformer}   & 0.494/0.479 & 0.538/0.504 & 0.574/0.521 & 0.562/0.535 \\
& TimesNet \cite{TimesNet}     & 0.384/0.402 & 0.436/0.429 & 0.491/0.469 & 0.521/0.500 \\
& LightTS \cite{LightTS}       & 0.424/0.432 & 0.475/0.462 & 0.518/0.488 & 0.547/0.533 \\
& \textbf{Ours}                & \textbf{0.366}/\textbf{0.397} & \textbf{0.420}/\textbf{0.426} & \textbf{0.458}/\textbf{0.455} & \textbf{0.503}/\textbf{0.496} \\
\hline

\multirow{7}{*}{ETTh2}
& Autoformer \cite{Autoformer} & 0.346/0.388 & 0.456/0.452 & 0.482/0.486 & 0.515/0.511 \\
& FEDformer \cite{FEDformer}   & 0.358/0.397 & 0.429/0.439 & 0.496/0.487 & 0.463/0.474 \\
& Stationary \cite{Stationary} & 0.476/0.458 & 0.512/0.493 & 0.552/0.551 & 0.562/0.560 \\
& ETSformer \cite{ETSformer}   & 0.340/0.391 & 0.430/0.439 & 0.485/0.479 & 0.500/0.497 \\
& TimesNet \cite{TimesNet}     & 0.340/0.374 & 0.402/0.414 & 0.452/0.452 & 0.462/0.468 \\
& LightTS \cite{LightTS}       & 0.397/0.437 & 0.520/0.504 & 0.626/0.559 & 0.863/0.672 \\
& \textbf{Ours}                & \textbf{0.337}/\textbf{0.368} & \textbf{0.399}/\textbf{0.409} & \textbf{0.437}/\textbf{0.443} & \textbf{0.451}/\textbf{0.463} \\
\hline

\multirow{7}{*}{ETTm1}
& Autoformer \cite{Autoformer} & 0.505/0.475 & 0.553/0.496 & 0.621/0.537 & 0.671/0.561 \\
& FEDformer \cite{FEDformer}   & 0.379/0.419 & 0.426/0.441 & 0.445/0.459 & 0.543/0.490 \\
& Stationary \cite{Stationary} & 0.386/0.398 & 0.459/0.444 & 0.495/0.464 & 0.585/0.516 \\
& ETSformer \cite{ETSformer}   & 0.375/0.398 & 0.408/0.410 & 0.435/0.428 & 0.499/0.462 \\
& TimesNet \cite{TimesNet}     & 0.338/0.375 & 0.374/0.387 & 0.410/\textbf{0.411} & 0.478/0.450 \\
& LightTS \cite{LightTS}       & 0.374/0.400 & 0.400/0.407 & 0.438/0.438 & 0.527/0.502 \\
& \textbf{Ours}                & \textbf{0.334}/\textbf{0.365} & \textbf{0.382}/\textbf{0.371} & \textbf{0.408}/\textbf{0.411} & \textbf{0.462}/\textbf{0.441} \\
\hline

\multirow{7}{*}{ETTm2}
& Autoformer \cite{Autoformer} & 0.255/0.339 & 0.281/0.340 & 0.339/0.372 & 0.433/0.432 \\
& FEDformer \cite{FEDformer}   & 0.203/0.287 & 0.269/0.328 & 0.325/0.366 & 0.421/0.415 \\
& Stationary \cite{Stationary} & 0.192/0.274 & 0.280/0.339 & 0.334/0.361 & 0.417/0.413 \\
& ETSformer \cite{ETSformer}   & 0.189/0.280 & 0.253/0.319 & 0.314/0.357 & 0.414/0.413 \\
& TimesNet \cite{TimesNet}     & 0.187/\textbf{0.267} & 0.249/0.309 & 0.321/0.351 & 0.408/0.403 \\
& LightTS \cite{LightTS}       & 0.209/0.308 & 0.311/0.382 & 0.442/0.466 & 0.675/0.587 \\
& \textbf{Ours}                & \textbf{0.176}/0.283 & \textbf{0.237}/\textbf{0.300} & \textbf{0.294}/\textbf{0.342} & \textbf{0.395}/\textbf{0.391} \\
\hline

\multirow{7}{*}{Traffic}
& Autoformer \cite{Autoformer} & 0.613/0.388 & 0.616/0.382 & 0.622/0.337 & 0.660/0.408 \\
& FEDformer \cite{FEDformer}   & 0.587/0.366 & 0.604/0.373 & 0.621/0.383 & 0.626/0.382 \\
& Stationary \cite{Stationary} & 0.612/0.338 & 0.613/0.340 & 0.618/\textbf{0.328} & 0.653/0.355 \\
& ETSformer \cite{ETSformer}   & 0.607/0.392 & 0.621/0.399 & 0.622/0.396 & 0.632/0.396 \\
& TimesNet \cite{TimesNet}     & 0.593/0.321 & 0.617/0.336 & 0.629/0.336 & 0.640/0.350 \\
& LightTS \cite{LightTS}       & 0.615/0.391 & 0.601/0.382 & 0.613/0.386 & 0.658/0.407 \\
& \textbf{Ours}                & \textbf{0.579}/\textbf{0.315} & \textbf{0.595}/\textbf{0.305} & \textbf{0.612}/\textbf{0.328} & \textbf{0.618}/\textbf{0.337} \\
\hline
\end{tabular}
\vspace{-5mm}
\end{table}

\subsection{Probabilistic Forecasting Performance}

We evaluate probabilistic accuracy using the Continuous Ranked Probability Score (CRPS) at a 24-step horizon. This short horizon setting is standard in probabilistic forecasting benchmarks \cite{LagLLaMA}, as it isolates distributional quality from long-range error accumulation. We compute CRPS by drawing 100 stochastic samples from the normalizing flow for each time step.

Table~\ref{tab:crps_ettm2} compares PaP-NF with native probabilistic baselines, including AutoETS, DynOptTheta, NPTS, CrostonSBA, and DeepAR. These baselines are drawn from the AutoGluon implementations \cite{GluonTS} to ensure consistent configuration across datasets. Deterministic long-term models are not included, as they lack native density estimation and require implementation-dependent modifications for CRPS evaluation.

\noindent PaP-NF achieves consistently competitive CRPS performance across all five datasets. It attains the best score on ETTh1 and matches the strongest baseline on ETTh2, while ranking second on ETTm1, ETTm2, and Traffic. This pattern demonstrates that PaP-NF not only maintains accuracy in stable regimes but also preserves stable uncertainty in more irregular settings.

\begin{table}[H]
\renewcommand{\arraystretch}{1.2} 
\centering
\setlength{\tabcolsep}{7pt}
\caption{CRPS comparison against \textbf{native probabilistic} baselines ($H=24$). The best results are in \textbf{bold}, and the second-best are \underline{underlined}.}
\label{tab:crps_ettm2}
\vspace{8pt}
\small
\begin{tabular}{l ccccc}
\toprule
\textbf{Model} & \textbf{ETTh1} & \textbf{ETTh2} & \textbf{ETTm1} & \textbf{ETTm2} & \textbf{Traffic} \\
\midrule
NPTS \cite{NPTS} & 0.268 & 0.216 & 0.162 & 0.139 & 0.191 \\
CrostonSBA \cite{Croston,Syntetos} & 0.123 & 0.112 & 0.094 & 0.102 & 0.414 \\
AutoETS \cite{AutoETS} & 0.117 & 0.105 & 0.073 & 0.081 & 0.492  \\
DynOptTheta \cite{DynOptTheta} & 0.117 & \underline{0.085} & \textbf{0.070} & \textbf{0.049} & 0.383 \\
DeepAR \cite{DeepAR} & \underline{0.105} & \textbf{0.082} & 0.074 & \underline{0.068} & \textbf{0.100}  \\
\midrule
\textbf{PaP-NF (Ours)} & \textbf{0.103} & \textbf{0.082} & \underline{0.071} & \underline{0.068} & \underline{0.181}  \\
\bottomrule
\end{tabular}
\end{table}

\subsection{Ablation Study}

We conduct ablation studies to evaluate the contributions of the PaP module, pre-trained LLM, global context, and prefix length. To avoid redundancy, we report representative settings that best highlight each component's impact.

\subsubsection{Effectiveness of Prefix-as-Prompt}
Table~\ref{tab:ablation_pap_vertical} quantifies the contribution of the PaP module to forecasting accuracy. Removing the PaP alignment (w/o PaP) leads to a significant MSE increase of 9.4\% at $H=720$, indicating a substantial loss in predictive stability. This performance drop is particularly pronounced as the horizon $H$ increases, confirming that the PaP structure is a requisite for handling long-range dependencies. These results empirically validate that the PaP module effectively bridges the internal representation of time series with the LLM backbone, a necessity that becomes more evident in extended forecasting tasks.

\begin{table}[t]
\centering
\caption{Ablation on PaP structure (ETTh1). Relative performance gap with respect to the ablated variant (w/o PaP) is reported in parentheses.}
\vspace{8pt}
\label{tab:ablation_pap_vertical}
\setlength{\tabcolsep}{10pt} 
\renewcommand{\arraystretch}{1.2}

\begin{tabular}{c c c c}
\hline
Horizon & Model & MSE & MAE \\
\hline \hline

\multirow{2}{*}{$H=96$}
  & w/o PaP            & 0.388 & 0.435 \\
  & \textbf{PaP-NF (Ours)} & \textbf{0.366 (-5.7\%)} & \textbf{0.417 (-4.1\%)} \\
\hline

\multirow{2}{*}{$H=720$}
  & w/o PaP            & 0.521 & 0.536 \\
  & \textbf{PaP-NF (Ours)} & \textbf{0.472 (-9.4\%)} & \textbf{0.495 (-7.7\%)} \\
\hline

\end{tabular}
\vspace{-3mm}
\end{table}

\subsubsection{Impact of Pre-trained Knowledge}
To verify the contribution of pre-trained knowledge in the LLM, we compare PaP-NF with and without a pre-trained backbone. In the w/o pre-trained LLM variant, the frozen LLaMA-3.1-8B backbone is replaced by a randomly initialized Transformer with the same architecture. All other components, including the PaP module and the normalizing flow decoder, remain unchanged. As shown in Table~\ref{tab:ablation_pre-trained}, the pre-trained LLM version achieves consistently lower MSE and MAE than the w/o pre-trained LLM variant across all horizons. The gap widens with the forecast horizon and is most pronounced at $H=720$, where MSE decreases by 7.8\% and MAE by 6.6\%.

\begin{table}[H]
\centering
\caption{Comparison of PaP-NF with and without a pre-trained LLM on ETTh1 ($H=720$).}
\setlength{\tabcolsep}{20pt}
\vspace{4pt}
\label{tab:ablation_pre-trained}
\begin{tabular}{c|cc}
\hline
\textbf{Model} & MSE & MAE \\
\hline
PaP-NF (w/o pre-trained LLM) & 0.498 & 0.487 \\
PaP-NF (pre-trained LLM) & \textbf{0.459} & \textbf{0.455} \\
\hline
\end{tabular}
\vspace{-3mm}
\end{table}

\subsubsection{Effect of Global Context}

To verify the contribution of the LLM-derived global context, we compare the full PaP-NF model with a variant that removes the global context branch. As shown in Table~\ref{tab:ablation_global_context}, removing the global context increases prediction errors on ETTh1 at $H=720$, with MSE rising from 0.459 to 0.481 and MAE from 0.455 to 0.477. These results indicate that the global context provides complementary long-range information beyond local numerical patterns, which is particularly important for long-horizon forecasting.

{\color{red}
\begin{table}[H]
\centering
\caption{Comparison of PaP-NF with and without global context on ETTh1 ($H=720$).}
\setlength{\tabcolsep}{20pt}
\vspace{4pt}
\label{tab:ablation_global_context}
\begin{tabular}{c|cc}
\hline
\textbf{Model} & MSE & MAE \\
\hline
PaP-NF (w/o global context) & 0.481 & 0.477 \\
PaP-NF (full) & \textbf{0.459} & \textbf{0.455} \\
\hline
\end{tabular}
\vspace{-7mm}
\end{table}
}

\subsubsection{Sensitivity to Prefix Length}

The performance variation with respect to the prefix length $K$ is analyzed in Table~\ref{tab:ablation_prefix}, where the best performance is observed at $K=5$ across all horizons. We observe a clear performance trade-off: a minimal length ($K=1$) provides insufficient alignment context, while excessively large values ($K=12$) introduce redundant information that disperses the LLM’s attention. Using $K=5$ gives the most consistent performance, suggesting that it provides enough context without adding unnecessary tokens to the attention computation. Such balance is crucial for maintaining the semantic stability of the PaP module in long-term forecasting tasks.

\begin{table}[H]
\centering
\caption{Sensitivity analysis of Prefix Length $K$ on ETTh1. $K=5$ yields the best trade-off.}
\vspace{5pt}
\setlength{\tabcolsep}{5pt}
\label{tab:ablation_prefix}
\resizebox{1.0\linewidth}{!}{
\begin{tabular}{c||cc|cc|cc|cc}
\hline
\multirow{2}{*}{Length ($K$)} & \multicolumn{2}{c|}{$\mathbf{H}=96$} & \multicolumn{2}{c|}{$\mathbf{H}=192$} & \multicolumn{2}{c|}{$\mathbf{H}=336$} & \multicolumn{2}{c}{$\mathbf{H}=720$} \\
\cline{2-9}
& MSE & MAE & MSE & MAE & MSE & MAE & MSE & MAE \\
\hline \hline
$K=1$ & 0.380 & 0.425 & 0.438 & 0.448 & 0.461 & 0.501 & 0.485 & 0.510 \\
$K=3$ & 0.370 & 0.419 & 0.428 & 0.438 & 0.449 & 0.491 & 0.478 & 0.501 \\
$\mathbf{K=5}$ \textbf{(Ours)} & \textbf{0.366} & \textbf{0.417} & \textbf{0.424} & \textbf{0.435} & \textbf{0.442} & \textbf{0.488} & \textbf{0.472} & \textbf{0.495} \\
$K=8$ & 0.371 & 0.421 & 0.430 & 0.440 & 0.450 & 0.495 & 0.480 & 0.505 \\
$K=12$ & 0.378 & 0.429 & 0.435 & 0.445 & 0.458 & 0.502 & 0.489 & 0.511 \\
\hline
\end{tabular}
}
\end{table}

\subsubsection{Model Efficiency}
A common concern with LLM-based forecasting is computational overhead. However, PaP-NF remains parameter-efficient by keeping the LLM backbone frozen and training only lightweight PaP and projection modules. Regarding inference latency, while probabilistic sampling naturally incurs a cost compared to deterministic point prediction, our Normalizing Flow decoder generates samples in a single forward pass, making it significantly faster than iterative diffusion-based models (e.g., CSDI \cite{CSDI}). This offers a favorable trade-off for risk-sensitive applications requiring robust uncertainty quantification.

\section{Discussion}

\begin{figure}[H]
    \centering
    \makebox[\linewidth][c]{%
        \includegraphics[width=1.0\linewidth]{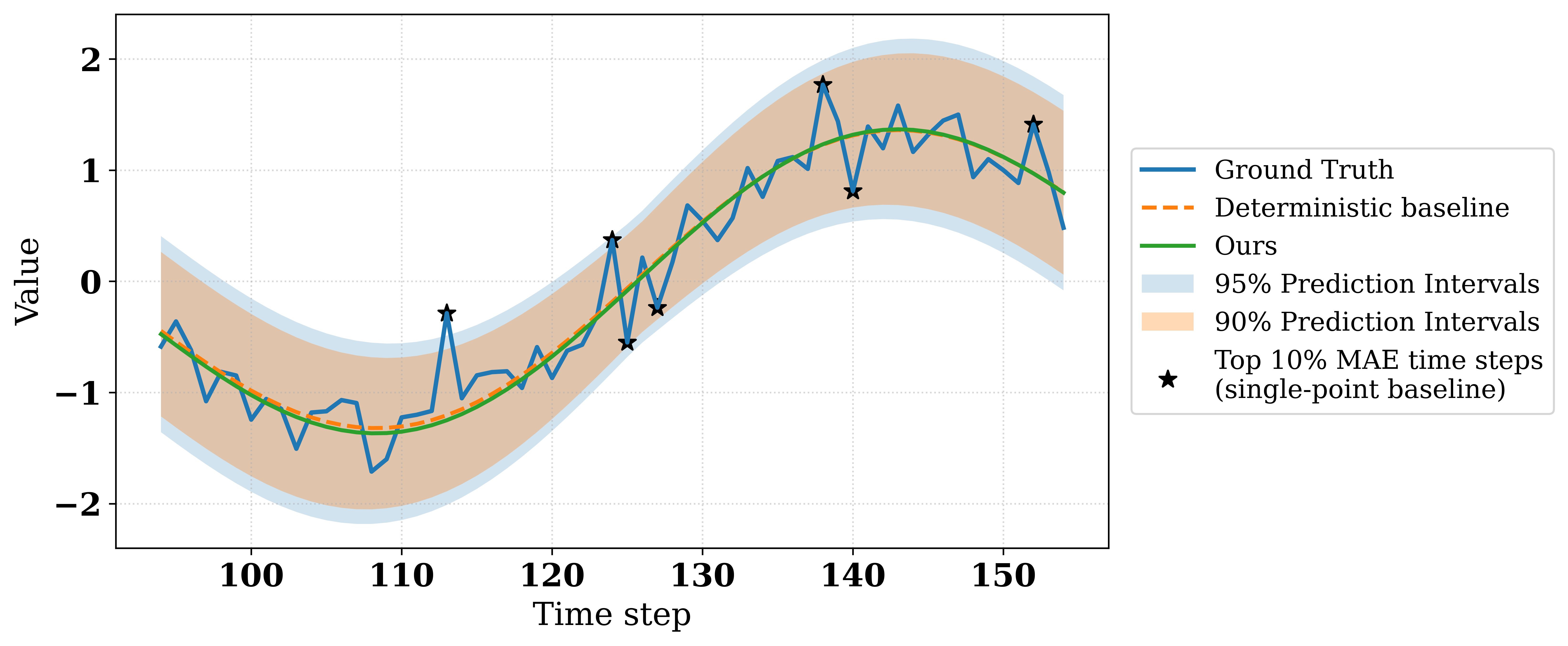}
    }
    \caption{Qualitative comparison on ETTm2 ($H=720$). PaP-NF generates prediction intervals versus deterministic point forecasts. Stars mark time steps where the deterministic baseline exhibits the highest absolute errors (top 10\%). PaP-NF captures these challenging points within its 90\% prediction intervals, illustrating its ability to model uncertainty and manage high-risk regions.}
    \label{fig:extreme_slice}
\vspace{-5mm}
\end{figure}

The empirical results in Table~\ref{tab:long_term_forecast_side_dataset} and Fig.~\ref{fig:extreme_slice} suggest the efficacy of decoupling global semantic reasoning from local numerical dynamics. Unlike deterministic baselines that collapse future uncertainty into a single, often erroneous trajectory, PaP-NF dynamically adjusts its predictive distribution via conditional normalizing flows. This capability helps the model remain more stable against the error accumulation typically seen in long-term horizons, as the pre-trained LLM provides a stable semantic reference that guides the generation process even under high volatility.

\noindent Further analysis through the ablation results shows that the model’s gains mainly come from the alignment mechanism and the use of pre-trained LLM representations. The degradation observed without the PaP module confirms its necessity as a bridge for the modality gap, while the underperformance of the untrained backbone proves that the model leverages the high-level contextual reasoning inherent in pre-trained weights rather than mere architectural depth.

\noindent In terms of computational cost and output quality, PaP-NF provides a practical balance relative to other generative approaches. While diffusion-based models like CSDI \cite{CSDI} suffer from $O(T)$ iterative sampling latency, our NF-based decoder achieves $O(1)$ efficiency in a single forward pass. Although the memory footprint of the LLaMA-3.1 backbone is larger than that of lightweight MLP-based models, the frozen parameter strategy ensures that training remains feasible on standard hardware. The additional memory cost is compensated by improved calibration and coverage.

\noindent\textbf{Qualitative Analysis of Uncertainty Coverage.}
Fig.~\ref{fig:extreme_slice} provides a qualitative comparison between PaP-NF and a deterministic baseline on ETTm2 for $H=720$. We use this setting because ETTm2 exhibits stronger non-stationarity than the hourly datasets, and $H=720$ represents the most challenging horizon among our benchmarks. We focus on time steps where the baseline exhibits the highest absolute errors (top 10\%), highlighted with star markers. These challenging points are largely covered by PaP-NF’s 90\% and 95\% prediction intervals. The model adapts the width of its uncertainty bands to local variation in the signal, maintaining coverage even when the underlying trajectory shifts rapidly. This shows that PaP-NF captures both central tendencies and distributional structure in a calibrated manner, supporting more reliable decisions in settings where risk and uncertainty must be explicitly managed.

\section{Conclusion}

We presented PaP-NF, a hybrid probabilistic framework that leverages pre-trained LLMs as global context encoders for long-term time series forecasting. By integrating Prefix-as-Prompt reprogramming with conditional normalizing flows, our model effectively captures complex, multi-modal future trajectories while bypassing the precision loss inherent in discrete tokenization. Extensive evaluations demonstrate that PaP-NF provides reasonable uncertainty quantification and maintains competitive point accuracy across diverse benchmarks.

\noindent The core insight of this study is that stable temporal reasoning is supported by the disjoint representation of numerical dynamics and semantic context. By preventing the collapse of heterogeneous temporal patterns into a single latent space, PaP-NF provides a practical direction for utilizing foundation models in time series analysis. Future research will focus on enhancing the deployability of this framework through backbone distillation and quantization, extending the reach of LLM-guided probabilistic reasoning to resource-constrained environments.

\begin{credits}
\subsubsection{\ackname}

This work was supported by the Inha University Research Grant.

\subsubsection{\discintname} The authors have no competing interests to declare that are relevant to the content of this article. 
\end{credits}

\end{document}